\pdfoutput=1

\documentclass[11pt]{article}

\usepackage{acl}

\usepackage{times}
\usepackage{latexsym}

\usepackage[T1]{fontenc}
\usepackage{textgreek}

\usepackage[utf8]{inputenc}

\usepackage{microtype}

\usepackage{subcaption}                     
\usepackage{graphicx}
\usepackage[nolist]{acronym}                
\usepackage{hyperref}                       
\usepackage{xurl}                           
\usepackage{booktabs}                       
\usepackage{multirow}                       
\usepackage{amsfonts}                       
\usepackage{nicefrac}                       
\usepackage{enumitem}                       
\usepackage{pgfplots}                       
\pgfplotsset{compat=1.17} 
\usepackage{pgfplotstable}

\title{An Empirical Study on Cross-X Transfer for Legal Judgment Prediction}

\author{
  Joel~Niklaus$^{\;\dagger\;}\thanks{\hspace{2mm}Equal contribution.}$
  \quad
    Matthias~Stürmer$^{\;\dagger}$
  \quad
  Ilias~Chalkidis$^{\;\ddagger\;\diamond\;*}$ \\
${\;\dagger}$ Institute of Computer Science, University of Bern, Switzerland \\
${\;\ddagger}$ Department of Computer Science, University of Copenhagen, Denmark \\
$^{\diamond\;}$Cognitiv+, Athens, Greece \\
}

\begin{document}
\maketitle

\begin{acronym}[UMLX]
    \acro{FSCS}{Federal Supreme Court of Switzerland}
    \acro{SCI}{Supreme Court of India}
    \acro{ECHR}{European Convention of Human Rights}
    \acro{ECtHR}{European Court of Human Rights}
    \acro{SCOTUS}{Supreme Court of the United States}
    \acro{SPC}{Supreme People's Court of China}
    \acro{SJP}{Swiss-Judgment-Prediction}
    \acro{ASO}{Almost Stochastic Order}
    \acro{ILDC}{Indian Legal Documents Corpus}
    
    \acro{US}{United States}
    \acro{EU}{European Union}

    \acro{NLP}{Natural Language Processing}
    \acro{ML}{Machine Learning}
    \acro{LJP}{Legal Judgment Prediction}
    \acro{SJP}{Swiss-Judgment-Prediction}
    
    \acro{BERT}{Bidirectional Encoder Representations from Transformers}

    \acro{CLT}{Cross-Lingual Transfer}
    \acro{HRL}{high resource language}
    \acro{LRL}{low resource language}
    \acro{NMT}{Neural Machine Translation}
    \acro{NLU}{Natural Language Understanding}
\end{acronym}

\begin{abstract}
Cross-lingual transfer learning has proven useful in a variety of Natural Language Processing (NLP) tasks, but it is understudied in the context of legal NLP, and not at all in Legal Judgment Prediction (LJP). We explore transfer learning techniques on LJP using the trilingual Swiss-Judgment-Prediction dataset, including cases written in three languages. We find that cross-lingual transfer improves the overall results across languages, especially when we use adapter-based fine-tuning. Finally, we further improve the model's performance by augmenting the training dataset with machine-translated versions of the original documents, using a $3\times$ larger training corpus. Further on, we perform an analysis exploring the effect of cross-domain and cross-regional transfer, i.e., train a model across domains (legal areas), or regions. We find that in both settings (legal areas, origin regions), models trained across all groups perform overall better, while they also have improved results in the worst-case scenarios. Finally, we report improved results when we ambitiously apply cross-jurisdiction transfer, where we further augment our dataset with Indian legal cases. 
\end{abstract}


\section{Introduction}
\label{sec:introduction}

Rapid development in \ac{CLT} has been achieved by pre-training transformer-based models in large multilingual corpora \cite{conneau_unsupervised_2020, xue-etal-2021-mt5}, where these models have state-of-the-art results in multilingual NLU benchmarks \cite{ruder-etal-2021-xtreme}.
Moreover, adapter-based fine-tuning \cite{houlsby2019parameterefficient, pfeiffer-etal-2020-mad} has been proposed to minimize the misalignment of multilingual knowledge (alignment) when \ac{CLT} is applied, especially in a zero-shot fashion, where the target language is unseen during training.
\ac{CLT} is severely understudied in legal \acs{NLP} applications except for \citet{chalkidis-etal-2021-multieurlex} who experimented with several methods for \ac{CLT} on MultiEURLEX, a newly introduced multilingual legal topic classification dataset, including EU laws. 

To the best of our knowledge, \ac{CLT} has not been applied to the \ac{LJP} task \cite{aletras_predicting_2016,xiao_cail2018_2018, chalkidis-etal-2019-neural,malik-etal-2021-ildc}, where the goal is to predict the verdict (court decision)  given  the  facts  of  a  legal  case. 
In this setting, positive impact of cross-lingual transfer is not as conceptually straight-forward as in other general applications (NLU), since there are known complications for sharing legal definitions and interpreting law across languages \cite{gotti-2014-translate,mcauliffe-2014-translate,robertson2016multilingual,ramos-2021-translate}.

\begin{figure}[t]
    \centering
    \resizebox{0.97\columnwidth}{!}{
    \includegraphics{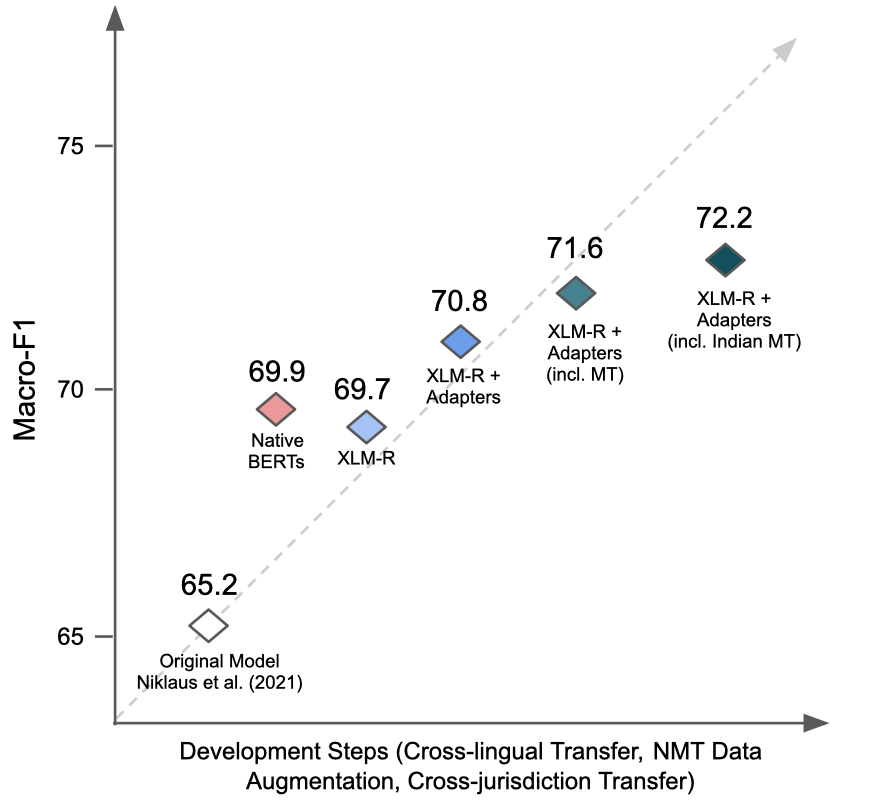}
    }
    \vspace{-2mm}
    \caption{
    Incremental performance improvement through several development steps.
    }.
    \label{fig:progress}
    \vspace{-10mm}
\end{figure}

Following the work of \citet{niklaus-etal-2021-swiss}, we experiment with their newly released trilingual \ac{SJP} dataset, containing cases from the \ac{FSCS}, written in three official Swiss languages (German, French, Italian). 
The dataset covers four legal areas (public, penal, civil, and social law) and lower courts located in eight regions of Switzerland (Zurich, Ticino, etc.), which poses interesting new challenges on model robustness / fairness and the effect of cross-domain and cross-regional knowledge sharing. In their experiments, \citet{niklaus-etal-2021-swiss} find that the performance in cases written in Italian is much lower compared to the rest, while also performance varies a lot across regions and legal areas.

\subsection*{Main Research Questions}
We pose and examine four main research questions:

\noindent\textbf{RQ1}: \emph{Is cross-lingual transfer beneficial across all or some of the languages?}

\noindent\textbf{RQ2}: \emph{Do models benefit or not from cross-regional and cross-domain transfer?}

\noindent\textbf{RQ3}: \emph{Can we leverage data from another jurisdiction to improve performance?}

\noindent\textbf{RQ4}: \emph{How does representational bias (wrt. language, origin region, legal area) affect model's performance?}

\subsection*{Contributions}
The contributions of this paper are fourfold:
\vspace{-2mm}

\begin{itemize}[leftmargin=8pt, itemsep=0em]
    \item We explore, for the first time, the application of cross-lingual transfer learning in the challenging \ac{LJP} task in several settings (Section~\ref{sec:cross_lingual_exps}). 
    We find that a pre-trained language model fine-tuned multilingually, outperforms its monolingual counterparts, especially when we use adapter-based fine-tuning and augment the training data with machine-translated versions of the original documents ($3\times$ larger training corpus) with larger gains in a low-resource setting (Italian). 
    \item  We perform cross-domain and cross-regional analyses (Section~\ref{sec:cross_domain_exps}) exploring the effects of cross-domain and cross-regional transfer, i.e., train a model across domains, i.e., legal areas (e.g., civil, penal law), or regions (e.g., Zurich, Ticino). We find that in both settings (legal areas, regions), models trained across all groups perform overall better and more robustly; while always improving performance in the worst-case (region or legal area) scenario.
    \item We also report improved results when we apply cross-jurisdiction transfer (Section~\ref{sec:cross_jurisdiction_exps}) , where we further augment our dataset with Indian legal cases originally written in English. 
    \item We release the augmented dataset (incl. 100K machine-translated documents) and our code for replicability and future experimentation.\footnote{\url{https://huggingface.co/datasets/swiss_judgment_prediction}}

\end{itemize}

The cumulative performance improvement amounts to 7\% overall and 16+\% in the low-resource Italian subset, compared to the best reported scores in \citet{niklaus-etal-2021-swiss}, while using cross-lingual and cross-jurisdiction transfer we improve for 2.3\% overall and 4.6\% for Italian over our strongest baseline (NativeBERTs).

\section{Dataset and Task description}
\subsection{Swiss Legal Judgment Prediction Dataset}
\label{sec:swiss_ljp}

We investigate the \ac{LJP} task on the Swiss-Judgment-Prediction (SJP) dataset \cite{niklaus-etal-2021-swiss}. The dataset contains 85K cases from the Federal Supreme Court of Switzerland (FSCS) from the years 2000 to 2020 written in German, French, and Italian. The court hears appeals focusing on small parts of the previous (lower court) decision, where they consider possible wrong reasoning by the lower court. The dataset provides labels for a simplified binary (\emph{approval}, \emph{dismissal}) classification task. Given the facts of the case, the goal is to predict if the plaintiff's request is valid or partially valid (i.e., the court \emph{approved} the complaint).  

Since the dataset contains rich metadata, such as legal areas and origin regions, we can conduct experiments on the robustness of the models (see Section \ref{sec:cross_domain_exps}). The dataset is not equally distributed; in fact, there is a notable representation disparity where Italian have far fewer documents (4K), compared to German (50K) and French (31K). Representation disparity is also vibrant with respect to legal areas and regions. We refer readers to the work of \citeauthor{niklaus-etal-2021-swiss} for detailed dataset statistics.

\subsection{Indian Legal Judgment Prediction Dataset}
\label{sec:indian_ljp}

The \ac{ILDC} dataset \cite{malik-etal-2021-ildc} comprises 30K cases from the Indian Supreme Court in English. The court hears appeals that usually include multiple petitions and rules a decision (\emph{accepted} vs. \emph{rejected}) per petition. Similarly to \citet{niklaus-etal-2021-swiss}, \citeauthor{malik-etal-2021-ildc} released a simplified version of the dataset
with binarized labels. In effect, the two datasets (SJP, ILDC) target the very same task (partial or full approval of plaintiff's claims), nonetheless in two different jurisdictions (Swiss Federation and India). Our main goal, when we use ILDC as a complement of SJP, is to assess the possibility of cross-jurisdiction transfer from Indian to Swiss cases (see Section~\ref{sec:cross_jurisdiction_exps}), an experimental scenario that has not been explored so far in the literature.

\subsection{NMT-based Data Augmentation}
\label{sec:easy_nmt}

In some of our experiments, we perform data augmentation using machine-translated versions of the original documents, i.e., translate a document originally written in a single language to the other two (e.g., from German to French and Italian). We performed the translations using the EasyNMT\footnote{\url{https://github.com/UKPLab/EasyNMT}} framework utilizing the \emph{many-to-many} \ac{NMT}  model 
of \citet{fan2020}.\footnote{The \emph{one-to-one} OPUS-MT \cite{tiedemann-thottingal-2020-opus} models did not have any model available from French to Italian (fr2it) at the time of the experiments.} A preliminary manual check of some translated samples showed sufficient translation quality to proceed forward. We release the machine-translated additional dataset for future consideration on cross-lingual experiments or quality assessment.

To the best of our knowledge, machine translation for data augmentation has not been studied in legal \ac{NLP}  applications, while it is generally a straight-forward, though under-studied idea. As we show in the experiments (see Section \ref{sec:cross_lingual_exps}), the translations are effective, leading to an average improvement of 1.6\% macro-F1 for standard fine-tuning and 0.8\% for adapter-based one (see Table \ref{tab:cross_lingual}). For the low-resource Italian subset, the improvement even amounts to 3.2\% and 1.6\%, respectively.

\begin{table*}[t]
\centering
    \resizebox{\textwidth}{!}{
    \begin{tabular}{lrc|ccc|cc}
    \toprule
    \bf Model & \bf \#D & \bf \#M   &             \bf German $\uparrow$ &  \bf French $\uparrow$ &   \bf Italian $\uparrow$ &  \bf All $\uparrow$ & (Diff. $\downarrow$)\\
    \midrule
    \multicolumn{8}{l}{A1. \emph{Monolingual: Fine-tune  on the \bf{tgt training set}} (\textbf{src} $=$ \textbf{tgt}) --- Baselines }  \\
    \midrule
    Prior SotA (\small \citeauthor{niklaus-etal-2021-swiss}) & 3-35K & N & 68.5 {\small ± 1.6} & 70.2 {\small ± 1.1} & 57.1 {\small ± 0.4} & 65.2 {\small ± 0.8} & ( 13.1 )\\
    \midrule
    NativeBERTs  & 3-35K & N & \underline{69.6} {\small ± 0.4} &  \underline{72.0} {\small ± 0.5} &   \underline{68.2} {\small ± 1.3} &  \underline{69.9} {\small ± 1.6} & ( 3.8 )\\
    XLM-R        & 3-35K & N &  68.2 {\small ± 0.3} &  69.9 {\small ± 1.6} &  65.9 {\small ± 1.2} &  68.0 {\small ± 2.0} & ( 4.0 )\\
    \midrule
    \multicolumn{8}{l}{A2. \emph{Monolingual: Fine-tune on the \bf{tgt training set incl. machine-translations}} (\textbf{src} $=$ \textbf{tgt}) } \\
    \midrule
     NativeBERTs   & 60K & N &  \underline{70.0} {\small ± 0.7} &  \underline{71.0} {\small ± 1.3} &   \underline{71.9} {\small ± 2.5} &  \underline{71.0} {\small ± 0.8} & ( 0.9 ) \\
    XLM-R       & 60K & N &  68.8 {\small ± 1.4} &  70.7 {\small ± 2.1} &   71.9 {\small ± 2.6} &  70.4 {\small ± 1.3} & ( 1.1 ) \\
    \midrule
    \multicolumn{8}{l}{B1. \emph{Cross-lingual: Fine-tune on \bf{all training sets}} (\textbf{src} $\subset$ \textbf{tgt})} \\
    \midrule
    XLM-R   & 60K & 1 &  68.9 {\small ± 0.3} &  71.1 {\small ± 0.3} & 68.9 {\small ± 1.4} &  69.7 {\small ± 1.0} & ( 2.2 ) \\
    XLM-R {\small+ Adapters} & 60K & 1 &  \underline{69.9} {\small ± 0.6} &  \underline{71.8} {\small ± 0.7} &  \underline{70.7} {\small ± 1.8} &  \underline{70.8} {\small ± 0.8} & ( 0.9 ) \\
    \midrule
    \multicolumn{8}{l}{B2. \emph{Cross-lingual: Fine-tune on \bf{all training sets incl. machine-translations}} (\textbf{src} $\subset$ \textbf{tgt})} \\
    \midrule
     XLM-R   & 180K & 1 &  70.2 {\small ± 0.5} &  71.5 {\small ± 1.1} &  72.1 {\small ± 1.2} &  71.3 {\small ± 0.7} & ( 1.9 ) \\
    XLM-R {\small+ Adapters} & 180K & 1 &  \underline{\textbf{70.3}} {\small ± 0.9} &  \underline{\textbf{72.1}} {\small ± 0.8} &  \underline{\textbf{72.3}} {\small ± 2.1} &  \underline{\textbf{71.6}} {\small ± 0.8} & ( 2.0 ) \\
    \midrule
    \multicolumn{8}{l}{C. \emph{Zero-shot Cross-lingual: Fine-tune on \bf{all training sets excl. tgt language}} (\textbf{src} $\neq$ \textbf{tgt})} \\
    \midrule    XLM-R   & 25-57K & 1 &  58.4 {\small ± 1.2} &  58.7 {\small ± 0.8} &  \underline{68.1} {\small ± 0.2} &  61.7 {\small ± 4.5} & ( 9.7 ) \\
    XLM-R {\small+ Adapters} & 25-57K & 1 &  \underline{62.5} {\small ± 0.6} &  \underline{58.8} {\small ± 1.5} &  67.5 {\small ± 2.2} &  \underline{62.8} {\small ± 3.7} & ( 8.7 ) \\
    \bottomrule
    \end{tabular}
     }
    \caption{Test results for all training set-ups (monolingual w/ or w/o translations, multilingual w/ or w/o translations, and zero-shot) w.r.t source (src) and target (tgt) language. Best overall results are in \textbf{bold}, and best per setting (group) are \underline{underlined}. 
    \#D is the number of training documents used. \#M is the number of models trained/used.
    The mean and standard deviation are computed across random seeds and across languages for the last column. Diff. shows the difference between the best and the worst performing language. \textbf{\emph{The adapter-based multilingually fine-tuned XLM-R model including machine-translated versions  ($3\times$ larger corpus) has the best overall results}}.}
    \label{tab:cross_lingual}
    \vspace{-4mm}
\end{table*}

\section{Experiments}
\label{sec:experiments}

\subsection{Hierarchical BERT}
\label{sec:hier_bert}
Since the examined dataset (\ac{SJP}) contains many documents with more than 512 tokens (90\% of the documents are up to 2048), we use Hierarchical BERT models \cite{chalkidis-etal-2019-neural,niklaus-etal-2021-swiss, dai-etal-2022-revisit} to encode up to 2048 tokens per document (4$\times$512 blocks). 

We split the text into consecutive blocks of 512 tokens and feed the first 4 blocks to a shared standard BERT encoder. Then, we aggregate the block-wise \texttt{CLS} tokens by passing them through another 2-layer transformer encoder, followed by max-pooling and a final classification layer. 

We re-use and expand the implementation released by \citet{niklaus-etal-2021-swiss},\footnote{\url{https://github.com/JoelNiklaus/SwissJudgementPrediction}} which is based on the Hugging Face library \cite{wolf_transformers_2020}. Notably, we first improve the masking of the blocks. Specifically, when the document has less than the maximum number (4) of blocks, we pad with extra sequences of \texttt{PAD} tokens, without the use of special tokens (\texttt{CLS}, \texttt{SEP}), as was previously performed. This minor technical improvement seems to affect the model's performance at large (group A1 Prior SotA vs. NativeBERTs –– Table~\ref{tab:cross_lingual}).

We experiment with monolingually pre-trained BERT models (aka NativeBERTs) and the multilingually pre-trained XLM-R
of \citet{conneau_unsupervised_2020}.
Specifically, for monolingual experiments (Native BERTs), we use German-BERT \cite{chan_deepset_2019} for German, 
CamemBERT \cite{martin_camembert_2020} for French, 
and UmBERTo \cite{parisi_umberto_2020} for Italian,
similar to \citet{niklaus-etal-2021-swiss}. 

In our multilingual experiments, we also assess the effectiveness of adapter-based fine-tuning \cite{houlsby2019parameterefficient, pfeiffer-etal-2020-mad}, in comparison to standard full fine-tuning. In this setting, adapter layers are placed after all feed-forward layers of XLM-R and are trained together with the parameters of the layer-normalization layers. The rest of the model parameters remain untouched. 

\subsection{Experimental Set Up}
\label{sec:experimental_setup}

We follow \citet{niklaus-etal-2021-swiss} and report macro-averaged F1 score to account for the high class-imbalance in the dataset (approx.\ 20/80 approval/dismissal ratio). We repeat each experiment with 3 different random seeds and report the average score and standard deviation across runs (seeds). We perform grid-search for the learning rate and report test results, selecting the hyper-parameters with the best development scores.\footnote{Additional details on model configuration, training, and hyper-parameter tuning can be found in Appendix~\ref{sec:hyperparam_tuning}.}

\subsection{Cross-lingual Transfer}
\label{sec:cross_lingual_exps}

We first examine \emph{cross-lingual transfer}, where the goal is to share (transfer) knowledge across languages, and we compare models in three main settings: 
(a) \emph{Monolingual} (see Section \ref{sec:monolingual}): fine-tuned per language, using either the documents originally written in the language, or an augmented training set including the machine-translated versions of all other documents (originally written in another language),
(b) \emph{Cross-lingual} (see Section \ref{sec:crosslingual}): fine-tuned across languages with or without the additional translated versions, and
(c) \emph{Zero-shot cross-lingual} (see Section \ref{sec:zeroshot_crosslingual}): fine-tuned across a subset of the languages excluding the target language at a time.
We present the results in Table~\ref{tab:cross_lingual}.

\subsubsection{Mono-Lingual Training}
\label{sec:monolingual}
We observe that the baseline of \emph{monolingually} pre-trained and fine-tuned models (NativeBERTs) have the best results compared to the \emph{multilingually} pre-trained but \emph{monolingually} fine-tuned XLM-R (group A1 -- Table~\ref{tab:cross_lingual}). Representational bias across languages (Section~\ref{sec:swiss_ljp}) seems to be a key part of performance disparity, considering the performance of the least represented language (Italian) compared to the rest (3K vs. 21-35K training documents). However, this is not generally applicable, i.e., French have better performance compared to German, despite having approx.\ 30\% less training documents. 

Translating the full training set provides a $3\times$ larger training set (approx.\ 180K in total) that ``equally'' represents all three languages.\footnote{Representational equality with respect to number of training documents per language, but possibly not considering text quality, since we use \ac{NMT}  to achieve that goal.} Augmenting the original training sets with translated versions of the documents (group A2 -- Table~\ref{tab:cross_lingual}), originally written in another language, improves performance in almost all (5/6) cases (languages per model). Interestingly, the performance improvement in Italian, which has the least documents (less than 1/10 compared to German), is the largest across languages with 3.7\% for NativeBERT (68.2 to 71.9) and 6\% for XLM-R (65.9 to 71.9) making Italian the best performing language after augmentation. Data augmentation seems more beneficial for XLM-R, which does not equally represent the three examined languages.\footnote{Refer to \citet{conneau_unsupervised_2020} for resources per language  used to pre-train XLM-R (50\% less tokens for Italian).}

\begin{table*}[t]
    \centering
    \resizebox{\textwidth}{!}{
    \begin{tabular}{lrl|cccccccc|c}
\toprule
\bf Origin Region & \bf \#D & \bf \#L &  \bf ZH &  \bf ES &   \bf CS &  \bf NWS  &   \bf EM &    \bf RL &   \bf TI &     \bf FED &  \bf All \\
\midrule
\multicolumn{12}{c}{Region-specific fine-tuning with MT data augmentation} \\
\midrule
Zürich (ZH)                     & 26.4K  & de        & \underline{65.5} &  65.6 &  63.7 &      68.2 &  62.0 &  57.9 &  63.2 &  54.8 &  62.6 \\
Eastern Switzerland (ES)        & 17.1K  & de        & 62.9 &  \underline{66.9} &  62.8 &      65.2 &  62.2 &  60.2 &  57.8 &  55.1 &  61.6 \\
Central Switzerland (CS)        & 14.4K  & de        & 62.5 &  65.5 &  \underline{63.2} &      65.1 &  60.7 &  57.8 &  60.5 &  55.9 &  61.4 \\
Northwestern Switzerland (NWS)  & 17.1K  & de        & 66.0 &  68.6 &  65.2 &      \underline{67.9} &  61.6 &  57.0 &  57.1 &  55.5 &  62.4 \\
Espace Mittelland (EM)          & 24.9K  & de,fr     & 64.1 &  66.6 &  63.3 &      66.7 &  \underline{64.0} &  66.8 &  63.2 &  58.4 &  64.1 \\
Région Lémanique (RL)           & 40.2K & fr,de     & 61.0 &  64.7 &  60.2 &      63.7 &  63.4 &  \underline{69.8} &  67.6 &  54.3 &  63.1 \\
Ticino (TI)                     & 6.9K  & it        & 55.0 &  56.3 &  53.2 &      54.5 &  56.0 &  54.7 &  \underline{66.0} &  53.1 &  56.1 \\
Federation (FED)                & 3.9K  & de,fr,it  & 57.5 &  59.6 &  56.8 &      58.9 &  55.0 &  56.5 &  53.5 &  \underline{54.9} &  56.6 \\
\midrule
\multicolumn{12}{c}{Cross-regional fine-tuning w/o MT data augmentation} \\
\midrule
XLM-R                           & 60K & de,fr,it & 68.5 &  71.3  &  67.7  &  71.2  &  69.0 &  71.4  &  67.4  &  64.6 &  68.9  \\
XLM-R {\small+ Adapters}        & 60K & de,fr,it & \textbf{69.2} &  \textbf{73.9}  &  67.9 &      72.6  &  69.0  &  \textbf{72.1} &  70.1  & 64.2  &  69.9  \\
\midrule
\multicolumn{12}{c}{Cross-regional fine-tuning with MT data augmentation} \\
\midrule
NativeBERTs                     & 180K & de,fr,it &  69.0 &  72.1 &   68.6 &      72.0 &   69.9 &   71.9 &  68.8 &  64.8 &  69.6 \\
XLM-R                           & 180K & de,fr,it & \bf 69.2 &  72.9 &  68.3 &       \bf 73.3 &    69.9 &  71.7 &   70.4 &   \bf 65.0 &   70.1 \\
XLM-R {\small+ Adapters} & 180K & de,fr,it & \textbf{69.2}  &  73.3  &  \textbf{69.9}  &      73.0 &  \textbf{70.3} &  \textbf{72.1}  &  \textbf{70.9} &  63.8  &  \textbf{70.3} \\
\bottomrule
\end{tabular}
}
    \vspace{-2mm}
    \caption{Test results for models trained per region or across all regions. Best overall results are in \textbf{bold}, and in-domain are \underline{underlined}.
    \#D is the total number of training examples. \#L are the languages covered.
    \textbf{\emph{Cross-regional transfer is beneficial for all regions and has the best overall results. The shared multilingual model trained across all languages and regions slightly outperforms the baseline (NativeBERTs).}}
    }
    \label{tab:cross_domain_origin_regions}
    \vspace{-4mm}
\end{table*}

\subsubsection{Cross-Lingual Training}
\label{sec:crosslingual}
We now turn to the \emph{cross-lingual transfer} setting, where we train XLM-R across all languages in parallel. We observe that cross-lingual transfer (group B1 -- Table~\ref{tab:cross_lingual}) improves performance (+4.5\% p.p.) across languages compared to the same model (XLM-R) fine-tuned in a monolingual setting (group A1 -- Table~\ref{tab:cross_lingual}). This finding suggests that cross-lingual transfer (and the inherited benefit of using larger multilingual corpora) has a significant impact, despite the legal complication of sharing legal definitions across languages.
Augmenting the original training sets with the documents translated across all languages, further improves performance (group B2 -- Table~\ref{tab:cross_lingual}).

\subsubsection{Zero-Shot Cross-Lingual Training}
\label{sec:zeroshot_crosslingual}
We also present results in a \emph{zero-shot cross-lingual} setting (group C -- Table~\ref{tab:cross_lingual}), where XLM-R is trained in two languages and evaluated in the third one (unseen in fine-tuning). We observe that German has the worst performance (approx. 10\% drop), which can be justified as German is a \emph{Germanic} language, while both French and Italian are  \emph{Romance} and share a larger part of the vocabulary. 

Contrarily, in case of Italian, the low-resource language in our experiments, the model strongly benefits from zero-shot cross-lingual transfer, leading to 2.2\% p.p. improvement, compared to the monolingually trained XLM-R. In other words, training XLM-R with much more (approx\ $20\times$) out-of-language (57K in German and French) data is better compared to training on the limited (3K) in-language (Italian) documents (68.1 vs. 65.9).

\subsubsection{Fine-tuning with Adapters}
\label{sec:adapters}
Across all cross-lingual settings (groups B-C -- Table~\ref{tab:cross_lingual}), the use of Adapters improves substantially the overall performance. The multilingual adapter-based XLM-R in group B1 (Table~\ref{tab:cross_lingual}) has comparable performance to the NativeBERTs models of group A2, where the training dataset has been artificially augmented with machine translations. In a similar setting (group B2 -- Table~\ref{tab:cross_lingual}),  the multilingual adapter-based XLM-R in group B2 has the best overall results, combining the benefits of both cross-lingual transfer and data augmentation. 

With respect to \emph{cross-lingual performance parity}, the adapter-based XLM-R model has also the highest performance parity (least diff. in the last column of Table~\ref{tab:cross_lingual}), while augmenting the dataset with \ac{NMT}  translations leads to both the worst-case (language) performance and best performance for the least represented language (Italian).

In conclusion, cross-lingual transfer with an augmented dataset comprised of the original and machine-translated versions of all documents, has the best overall performance with a vibrant improvement (3\% compared to our strong baselines -- second part of Group A1 in Table~\ref{tab:cross_lingual}) in Italian, the least represented language.

\begin{table*}[ht]
\centering
    \resizebox{\textwidth}{!}{
    \begin{tabular}{lr|cccc|c}
    \toprule
    \bf Legal Area & \bf \#D &                 \bf Public Law &                   \bf Civil Law &                    \bf Penal Law &                  \bf Social Law &                      \bf All \\
    \midrule
    \multicolumn{7}{c}{Domain-specific fine-tuning with MT data augmentation} \\
    \midrule
    Public Law & 45.6K & \underline{56.4} {\small ± 2.2} &  52.2 {\small ± 2.0} &   59.7 {\small ± 4.9} &  60.1 {\small ± 5.8} &   57.1 {\small ± 3.2} \\
    Civil Law  & 34.5K & 44.4 {\small ± 7.9} &  \underline{64.2} {\small ± 0.6} &  45.5 {\small ± 13.1} &  43.6 {\small ± 5.2} &   49.4 {\small ± 8.6} \\
    Penal Law  & 35.4K & 40.8 {\small ± 10.1} &  55.8 {\small ± 2.9} &   \textbf{\underline{84.5}} {\small ± 1.3} &  61.1 {\small ± 7.5} &  60.6 {\small ± 15.7} \\
    Social Law & 29.1K  & 52.6 {\small ± 4.2} &  56.6 {\small ± 2.0} &   69.0 {\small ± 5.5} &  \underline{70.2} {\small ± 2.0} &   62.1 {\small ± 7.6} \\
    \midrule
    \multicolumn{7}{c}{Cross-domain fine-tuning w/o MT data augmentation} \\
    \midrule
    XLM-R & 60K & 57.4 {\small ± 2.0} &  66.1 {\small ± 3.1} &   81.4 {\small ± 1.4} &  70.8 {\small ± 2.0}  &   68.9 {\small ± 8.7} \\
     XLM-R {\small+ Adapters} & 60K &   58.4 {\small ± 2.5} &  66.1 {\small ± 2.4} &  83.1 {\small ± 1.2} &  71.1 {\small ± 1.4} &   69.7 {\small ± 9.0} \\
    \midrule
    \multicolumn{7}{c}{Cross-domain fine-tuning with MT data augmentation} \\
    \midrule
    NativeBERTs & 180K & 58.1 {\small ± 3.0} &  64.5 {\small ± 3.7} &   83.0 {\small ± 1.3} & 71.1 {\small ± 4.3} &   69.2 {\small ± 9.2} \\
     XLM-R         & 180K &   58.0 {\small ± 3.0} &  \textbf{67.2} {\small ± 1.6} &   84.4 {\small ± 0.2} &  70.2 {\small ± 1.3} &   \textbf{70.0} {\small ± 9.5} \\
     XLM-R {\small+ Adapters}         &  180K &   \textbf{58.6} {\small ± 2.7} &  66.8 {\small ± 2.8} &  83.1 {\small ± 1.3} &  \textbf{71.3} {\small ± 2.4} & 69.9 {\small ± 8.8} \\
    \bottomrule
    \end{tabular}
    }
    \caption{Test results for models (XLM-R with MT unless otherwise specified) \textbf{fine-tuned} per legal area (domain) or across all legal areas (domains). Best overall results are in \textbf{bold}, and in-domain are \underline{underlined}. The mean and standard deviations are computed across languages per legal area and across legal areas for the right-most column. \#D is the total number of training examples. \textbf{\emph{Cross-domain transfer is beneficial for 3 out of 4 legal areas and has the best overall results.}} The shared multilingual model trained across all languages and legal areas outperforms the baseline (monolingual BERT models).}
    \label{tab:cross_domain_legal_areas}
    \vspace{-4mm}
\end{table*}

\subsection{Cross-Domain/Regional Transfer Analysis}
\label{sec:cross_domain_exps}

Further on, we examine the benefits of transfer learning (knowledge sharing) in other dimensions. Hence, we analyze model performance with respect to origin regions and legal areas (domains of law).

\subsubsection{Origin Regions}
\label{sec:cross_domain_origin_regions}

In Table \ref{tab:cross_domain_origin_regions} we present the results for \emph{cross-regional} transfer.
In the top section of the table, we present results with region-specific multilingual (XLM-R) models evaluated across regions (in-region on the diagonal, zero-shot otherwise). We observe that the cross-regional models (two lower groups of Table~\ref{tab:cross_domain_origin_regions}) always outperform the region-specific models. Moreover, cross-lingual transfer is beneficial across cases, while adapter-based fine-tuning further improves results in 5 out of 8 cases (regions). Data augmentation is also beneficial in most cases. 

In the top part of Table~\ref{tab:cross_domain_origin_regions}, in 60\% of the cases (regions: ZH, ES, CS, NWS, TI), a ``zero-shot'' model, i.e., trained in the cases of another region, slightly outperforms the in-region model.
In other words, in almost every case (target region), there is another \emph{monolingual} region-specific model that outperforms the in-region one.

We consider two main factors that may explain these results: (a) the region-wise \emph{representational bias} considering the number of cases per region, and (b) the cross-regional \emph{topical similarity} of the training and test subsets across different regions. To approximate the cross-regional topical similarity, we consider the distributional similarity (or dissimilarity) w.r.t.\ legal areas (Table~\ref{tab:legal_area_distribution_distances} in Appendix~\ref{sec:distances}). None of these factors can fully explain the results. Although in 3 out of 5 cases, the best performing (out-of-region) model has been trained on more data compared to the in-region one. There are also other confounding factors (e.g., language), i.e., models trained on the cases of either Espace Mittelland (EM) or Région Lémanique (RL), both bilingual with 8-10K cases, have the best results across all single-region models, hence a further exploration of the overall dynamics is needed.


\begin{table*}[ht]
\centering
    \resizebox{\textwidth}{!}{
    \begin{tabular}{llr|ccc|cc}
    \toprule
    \bf Model & \bf Training Dataset & \bf \#D   &   \bf German $\uparrow$   &  \bf French $\uparrow$ & \bf Italian $\uparrow$ &  \bf All & (Diff. $\downarrow$)\\
    \midrule
    \multicolumn{7}{c}{Cross-lingual fine-tuning w/ or w/o MT data augmentation} \\
    \midrule
     XLM-R   & Original & 60K &  68.9 {\small ± 0.3} &  71.1 {\small ± 0.3} & 68.9 {\small ± 1.4} &  69.7 {\small ± 1.0} & ( 2.2 )\\
    XLM-R {\small+ Adapters} & Original & 60K &  \underline{69.9} {\small ± 0.6} &  \underline{71.8} {\small ± 0.7} &  \underline{70.7} {\small ± 1.8} &  \underline{70.8} {\small ± 0.8}  & ( 0.9 ) \\
    \midrule
    XLM-R &  + MT Swiss     & 180K                & 70.2 {\small ± 0.5} &  71.5 {\small ± 1.1} & \underline{72.1} {\small ± 1.2} & 71.3 {\small ± 0.7}  & ( 1.9 ) \\
    XLM-R {\small+ Adapters} & + MT Swiss  & 180K  & \underline{70.3} {\small ± 0.8} &  \underline{72.1} {\small ± 0.8} & \underline{72.1} {\small ± 1.2} & \underline{71.5} {\small ± 0.9} & ( 1.8 ) \\
    \midrule
    \multicolumn{7}{c}{Cross-jurisdiction fine-tuning w/ MT data augmentation} \\
    \midrule    
    XLM-R &  + MT \{Swiss, Indian\} & 276K               & 70.5 {\small ± 0.4} & 71.8 {\small ± 0.3} & \textbf{\underline{73.5}} {\small ± 1.4} & 72.0 {\small ± 0.9}  & ( 3.0 ) \\
    XLM-R {\small+ Adapters} & + MT \{Swiss, Indian\}   & 276K     & \textbf{\underline{71.0}} {\small ± 0.4}  & \textbf{\underline{73.0}} {\small ± 0.6} & 72.6 {\small ± 1.1} & \textbf{72.2} {\small ± 1.2}  & ( 2.0 )\\
    \midrule
    \multicolumn{7}{c}{Cross-jurisdiction zero-shot fine-tuning w/ MT data augmentation} \\
    \midrule  
    XLM-R & MT Indian         & 96K             & 50.4 {\small ± 1.5} & 47.9 {\small ± 1.0} & 49.5 {\small ± 1.3} & 49.3 {\small ± 1.0}  & ( 2.5)  \\
    XLM-R {\small+ Adapters} & MT Indian    & 96K   & \underline{51.6} {\small ± 2.9} & \underline{49.7} {\small ± 1.4} & \underline{50.1} {\small ± 1.4} & \underline{50.5} {\small ± 1.0}  & ( 1.9 ) \\
    \bottomrule
    \end{tabular}
    }
    \caption{Test results for cross-jurisdiction transfer. We present results in four settings: \emph{standard} (Original)  \emph{augmented} (+ MT Swiss), \emph{further augmented incl. cross-jurisdiction} (+ MT Swiss + MT Indian) and \emph{zero-shot} (MT Indian). Best results are in \textbf{bold}. Diff. shows the difference between the best performing language and the worst performing language (max - min). \textbf{\emph{Further augmenting with translated Indian cases is overall beneficial.}}
    }
    \label{tab:cross_jur}
    \vspace{-4mm}
\end{table*}

\subsubsection{Legal Areas}
\label{sec:cross_domain_legal_areas}
In Table \ref{tab:cross_domain_legal_areas} we present the results for \emph{cross-domain} transfer between legal areas (domains of law). 
The results on the diagonal (\underline{underlined}) are in-domain, i.e., fine-tuned and evaluated in the same legal area. We observe that for each domain, the models trained on in-domain data have the best results in the respective domain compared to the rest. 

Interesting to note is that the best results (\textbf{bold}) are achieved in the cross-domain setting in 3 out of 4 legal areas. Such an outcome is not anticipated based on the current trends in law industry, where legal experts (judges, lawyers) over-specialize and excel in specific legal areas, e.g., criminal defense lawyers. Penal law poses the only exception where the domain-specific model is on par with the cross-domain model. Again, the results per area do not correlate with the volume of training data (\emph{cross-domain representational bias}), and suggest that other qualitative characteristics (e.g., the idiosyncrasies of criminal law) affect the task complexity.

Similarly to the cross-regional experiments, the shared multilingual model (XLM-R) trained across all languages and legal areas with an augmented dataset outperforms the NativeBERTs models trained in a similar setting, giving another indication that the performance gains from cross-lingual transfer and data augmentation via machine translation are robust across domains as well.

\subsection{Cross-Jurisdiction Transfer}
\label{sec:cross_jurisdiction_exps}

We, finally, ``ambitiously'' stretch the limits of transfer learning in \ac{LJP} and we apply \emph{cross-jurisdiction} transfer, i.e., use of cases from different legal systems, another form of cross-domain transfer. For this purpose, we further augment the SJP dataset of \ac{FSCS} cases, with cases from the \ac{SCI}, published by \citet{malik-etal-2021-ildc}.\footnote{Although the \ac{SCI} rules under the Indian jurisdiction (law), while the \ac{FSCS} under the Swiss one, we hypothesize that the fundamentals of law in two modern legal systems are quite common and thus transferring knowledge could potentially have a positive effect. We discuss this matter in Section~\ref{sec:cross_jur_motivation}.} We consider and translate all (approx.\ 30K) Indian cases ruled up to the last year (2014) of our training dataset, originally written in English, to all target languages (German, French, and Italian).\footnote{We do not use the original documents written in English, as English is not one of our target languages.} 

In Table~\ref{tab:cross_jur}, we present the results for two cross-jurisdiction settings: \emph{zero-shot}  (Only MT Indian), where we train XLM-R on the machine-translated version of Indian cases, and \emph{further augmented} (Original + MT Swiss + MT Indian), where we further augment the (already augmented) training set of Swiss cases with the translated Indian ones. While zero-shot transfer clearly fails; interestingly, we observe improvement for all languages in the further augmented setting. This opens a fascinating new direction for \ac{LJP} research.

Similar to our results in Section~\ref{sec:cross_lingual_exps} with respect to cross-lingual performance parity, the standard adapter-based XLM-R model has also the highest performance parity (least diff. on Table~\ref{tab:cross_jur}), while the same model trained on the fully augmented dataset leads to the worst-case (language; German) performance and best performance for the least represented language (Italian).
\vspace{3mm}

\noindent The cumulative improvement from all applied enhancements adds up to 7\% macro-F1 compared to the XLM-R baseline and 16\% to the best method by \citet{niklaus-etal-2021-swiss} in the low-resource Italian subset, while using cross-lingual and cross-jurisdiction transfer we improve for 2.3\% overall and 4.6\% for Italian over our strongest baseline (NativeBERTs). 

Since our experiments present several incremental improvements, we assess the stability of the performance improvements with statistical significance testing by comparing the most crucial settings in Appendix \ref{sec:statistical}.

\section{Related Work}
\label{sec:related_work}

\paragraph{Legal Judgment Prediction}
(\ac{LJP}) is the task, where given the facts of a legal case, a system has to predict the correct outcome (legal judgement).
 Many prior works experimented with some forms of \ac{LJP}, 
however, the precise formulation of the \ac{LJP} task is non-standard as the jurisdictions and legal frameworks vary. 
\citet{aletras_predicting_2016,Medvedeva2018, chalkidis-etal-2019-neural} predict the plausible violation of \ac{ECHR} articles of the \ac{ECtHR}. \citet{xiao_cail2018_2018,xiao2021} study Chinese criminal cases where the goal is to predict the ruled duration of prison sentences and/or the relevant law articles. 

Another setup is followed by \citet{sulea_predicting_2017,malik-etal-2021-ildc,niklaus-etal-2021-swiss}, which use cases from Supreme Courts (French, Indian, Swiss, respectively), hearing appeals from lower courts relevant to several fields of law (legal areas). Across tasks (datasets), the goal is to predict the binary verdict of the court (approval or dismissal of the examined appeal) given a textual description of the case. 
None of these works have explored neither cross-lingual nor cross-jurisdiction transfer, while the effects of cross-domain and cross-regional transfer are also not studied. 

\paragraph{Cross-Lingual Transfer}

(\ac{CLT}) is a flourishing topic with the application of pre-trained transformer-based models trained in a multilingual setting \cite{devlin_bert_2019, lample_conneau_2019,conneau_unsupervised_2020, xue-etal-2021-mt5} excelling in NLU benchmarks \cite{ruder-etal-2021-xtreme}. Adapter-based fine-tuning \cite{houlsby2019parameterefficient,pfeiffer-etal-2021-adapterfusion} has been proposed as an anti-measure to mitigate misalignment of multilingual knowledge when \ac{CLT} is applied, especially in a zero-shot fashion, where the target language is unseen during training (or even pre-training).


Meanwhile, \ac{CLT} is understudied in legal \ac{NLP}  applications. \citet{chalkidis-etal-2021-multieurlex} experiment with standard fine-tuning, while they also examined the use of adapters \cite{houlsby2019parameterefficient} for zero-shot \ac{CLT} on a legal topic classification dataset comprising \ac{EU} laws. They found adapters to achieve the best tradeoff between effectiveness and efficiency. Their work did not examine the use of methods incorporating translated versions of the original documents in any form, i.e., translate train documents or test ones. Recently, \citet{xenouleas-etal-2022} used an updated, unparalleled version of \citeauthor{chalkidis-etal-2021-multieurlex} dataset to study \ac{NMT} -augmented CLT methods.  Other multilingual legal \ac{NLP}  resources \cite{galassi-etal-2020-cross,drawzeski-etal-2021-corpus} have been recently released, although \ac{CLT} is not applied in any form.

\section{Motivation and Challenges for Cross-Jurisdiction Transfer}
\label{sec:cross_jur_motivation}

Legal systems vary from country to country. Although they develop in different ways, legal systems also have some similarities based on historically accepted justice ideals, i.e., the rule of law and human rights. Switzerland has a civil law legal system \cite{walther_2001}, i.e., statutes (legislation) is the primary source of law, at the crossroads between Germanic and French legal traditions.

Contrary, India has a hybrid legal system with a mixture of civil, common law, i.e., judicial decisions have precedential value, and customary, i.e., Islamic ethics, or religious law \cite{indian-law-reuters}. The legal and judicial system derives largely from the British common law system, coming as a consequence of the British colonial era (1858-1947) \cite{singh-etal-2019}.

Based on the aforementioned, cross-jurisdiction transfer is challenging since the data (judgments) abide to different law standards.
Although the Supreme Court of India (SCI) rules under the Indian jurisdiction (law), while the Federal Supreme Court of Switzerland (FSCS) under the Swiss one, we hypothesize that the fundamentals of law in two modern legal systems are quite common and thus transferring knowledge could potentially have a positive effect, and thus it is an experiment worth considering, while we acknowledge that from a legal perspective equating legal systems is deeply problematic, since the legislation, the case law, and legal practice are different.
 
Our empirical work and experimental results shows that cross-jurisdiction transfer in this specific setting (combination of Swiss and Indian decisions) has a positive impact in performance, but we cannot provide any profound hypothesis neither we are able to derive any conclusions on the importance of this finding on legal literature and practice. We leave these questions in the hands of those who can responsibly bear the burden, the legal scholars.

\section{Conclusions and Future Work}
\label{sec:conclusions_and_future_work}

\subsection{Answers to the Research Questions}
Following the experimental results (Section~\ref{sec:experiments}), we answer the original predefined research questions:
\vspace{1mm}

\noindent\textbf{RQ1}: \emph{Is cross-lingual transfer beneficial across all or some of the languages?}
In Section~\ref{sec:cross_lingual_exps}, we find that vanilla \ac{CLT} is beneficial in a low-resource setting (Italian), with comparable results in the rest of the languages. Moreover, \ac{CLT} leveraging \ac{NMT} -based data augmentation is beneficial across all languages. Overall, our experiments lead to a single multi-lingual cross-lingually ``fairer'' model.\vspace{1mm}

\noindent\textbf{RQ2}: \emph{Do models benefit or not from cross-regional and cross-domain transfer?}
In Section~\ref{sec:cross_domain_exps}, we find that models benefit from cross-regional transfer across all cases, since they are exposed to (trained in) many more documents (cases). We believe cross-regional diversity is not a significant aspect, compared to the importance of the increased data volume and language diversity. Cross-domain transfer is beneficial in three out of four cases (legal areas), with comparable results on penal (criminal) law, where the application of law seems to be more straight-forward / standardized (higher performing legal area). Cross-regional and cross-domain transfer lead to more robust models.\vspace{1mm}

\noindent\textbf{RQ3}: \emph{Can we leverage data from another jurisdiction to improve performance?}
In Section~\ref{sec:cross_jurisdiction_exps}, we find that cross-jurisdiction transfer in our specific setup, i.e.,  very similar \ac{LJP} tasks, is beneficial. Again, we believe that this is mostly a matter of additional unique data (cases), rather than a matter of jurisdictional similarity. Cross-jurisdiction transfer leads to a better performing model.\vspace{1mm}

\noindent\textbf{RQ4}: \emph{How does representational bias (wrt. language, origin region, legal area) affect model's performance?}
We observe that representational bias -- in non-extreme cases (e.g., w.r.t. language) -- does not always explain performance disparities across languages, regions, or domains, and other characteristics also need to be considered. 

\subsection{Conclusions - Summary}
We examined the application of \acf{CLT} in \acf{LJP} for the very first time, finding a multilingually trained model to be superior when augmenting the dataset with \ac{NMT}. 
Adapter-based fine-tuning leads to even better results. 
We also examined the effects of cross-domain (legal areas) and cross-regional transfer, which is overall beneficial in both settings, leading to more robust models. 
Cross-jurisdiction transfer by augmenting the training set with machine-translated Indian cases further improves performance. 

\subsection{Future Work}
In future work, we would like to explore the use of a legal-oriented multilingual pre-trained model by either continued pre-training of XLM-R, or pre-training from scratch in multilingual legal corpora. Legal \ac{NLP}  literature \cite{chalkidis-etal-2022-lexglue, Zheng2021} suggests that domain-specific language models positively affect performance. 

In another interesting direction, we will consider other data augmentation techniques \cite{feng-etal-2021-survey, ma2019nlpaug} that rely on textual alternations (e.g., paraphrasing, etc.). We would also like to further investigate cross-jurisdictional transfer, either exploiting data for similar \ac{LJP}  tasks, or via multi-task learning on multiple \ac{LJP}  datasets with dissimilar task specifications.


\vspace{-2mm}
\section{Ethics Statement}
\vspace{-2mm}
The scope of this work is to study \ac{LJP} to broaden the discussion and help practitioners to build assisting technology for legal professionals and laypersons. We believe that this is an important application field, where research should be conducted \cite{tsarapatsanis-aletras-2021-ethical} to improve legal services and democratize law, while also highlight (inform the audience on) the various multi-aspect shortcomings seeking a responsible and ethical (fair) deployment of legal-oriented technologies.

In this direction, we study how we could better exploit all the available resources (from various languages, domains, regions, or even different jurisdictions). This combination leads to models that improve overall performance -- more robust models --, while having improved performance in the worst-case scenarios across many important demographic or legal dimensions (low-resource language, worst performing legal area and region).

Nonetheless, irresponsible use (deployment) of such technology is a plausible risk, as in any other application (e.g., online content moderation) and domain (e.g., medical). We believe that similar technologies should only be deployed to assist human experts (e.g., legal scholars in research, or legal professionals in forecasting or assessing legal case complexity) with notices on their limitations. 

The main examined dataset, \acf{SJP}, released by \citet{niklaus-etal-2021-swiss}, comprises publicly available cases from the \ac{FSCS}, where cases are pre-anonymized, i.e., names and other sensitive information are redacted. The same applies for the second one, \acf{ILDC} of \citet{malik-etal-2021-ildc}.

\section*{Acknowledgements}
This work has been supported by the Swiss National Research Program “Digital Transformation” (NRP-77)\footnote{\url{https://www.nfp77.ch/en/}} grant number 187477. This work is also partly funded by the Innovation Fund Denmark (IFD)\footnote{\url{https://innovationsfonden.dk/en}} under File No.\ 0175-00011A.  This research has been also co‐financed by the European Regional Development Fund of the European Union and Greek national funds through the Operational Program Competitiveness, Entrepreneurship and Innovation, under the call RESEARCH – CREATE – INNOVATE (Τ2ΕΔΚ-03849). 

We would like to thank Thomas Lüthi for his legal advice, Mara Häusler for great discussions regarding the evaluation process of the models, and Phillip Rust and Desmond Elliott for providing valuable feedback on the original draft of the manuscript.


\bibliography{anthology,custom}
\bibliographystyle{acl_natbib}

\appendix

\section{Hyperparameter Tuning}
\label{sec:hyperparam_tuning}
We experimented with learning rates in \{1e-5, 2e-5, 3e-5, 4e-5, 5e-5\} as suggested by \citet{devlin_bert_2019}. However, like reported by \citet{mosbach2020stability}, we also found RoBERTa-based models to exhibit large training instability with learning rate 3e-5, although this learning rate worked well for BERT-based models. 1e-5 worked well enough for all models.  To avoid either over- or under-fitting, we use Early Stopping \cite{caruana_overfitting_2001} on development data.
To combat the high class imbalance, we use oversampling, following \cite{niklaus-etal-2021-swiss}.

We opted to use the standard Adapters of \citet{houlsby2019parameterefficient}, as the language Adapters introduced by \citet{pfeiffer-etal-2020-mad} are more resource-intensive and require further pre-training per language. We tuned the adapter reduction factor in \{2$\times$, 4$\times$, 8$\times$, 16$\times$\} and got the best results with 2$\times$ and 4$\times$; we chose 4$\times$ for the final experiments to favor less additional parameters. We tuned the learning rate in \{1e-5, 5e-5, 1e-4, 5e-4, 1e-3\} and achieved the best results with 5e-5.

We additionally applied label smoothing \cite{szegedy-etal-2015-label} on cross-entropy loss. We achieved the best results with a label smoothing factor of 0.1 after tuning with \{0, 0.1, 0.2, 0.3\}.

\begin{table}[h]
\centering
\resizebox{\columnwidth}{!}{
\begin{tabular}{lrrrr}
\toprule
Model Type              &  M1 & M2 &  M3 &  M4 \\
\midrule
M1: NativeBERTs          &            1.0 &           1.0 &             1.0 &                1.0 \\
M2: NativeBERTs + MT CH  &            0.0 &           1.0 &             1.0 &                1.0 \\
M3: XLM-R + MT CH       &            0.0 &           0.0 &             1.0 &                1.0 \\
M4: XLM-R + MT CH + IN  &            0.0 &           0.0 &             0.0 &                1.0 \\
\bottomrule
\end{tabular}
}
\caption{Almost stochastic  dominance ($\epsilon_\mathrm{min} < 0.5$) with ASO. \emph{+ MT CH} stands for augmentation with machine translation inside the Swiss dataset and \emph{+ MT CH+IN} is the code for augmentation with machine-translations with the Swiss \textbf{and} Indian dataset.}
\label{tab:aso_scores}
\vspace{-3mm}
\end{table}

\section{Statistical Significance Testing}
\label{sec:statistical}
Since our experiments present several incremental improvements, we assessed the stability of the performance improvements with statistical significance testing by comparing the most crucial settings. Using \ac{ASO} \cite{dror-etal-2019-deep} with a confidence level $\alpha\!=\!0.05$, we find the score distributions of the core models (NativeBERTs, w/ and w/o MT Swiss, XLM-R w/ and w/o MT Indian and/or Swiss) stochastically dominant ($\epsilon_\mathrm{min} = 0$) over each other in order. 
We compared all pairs of models based on three random seeds each using ASO with a confidence level of  $\alpha = 0.05$ (before adjusting for all pair-wise comparisons using the Bonferroni correction). Almost stochastic  dominance ($\epsilon_\mathrm{min} < 0.5$) is indicated in Table~\ref{tab:aso_scores} in Appendix~\ref{sec:hyperparam_tuning}. We use the deep-significance Python library of \citet{dennis_ulmer_2021_4638709}.

\section{Distances Between Legal Area Distributions per Origin Regions}
\label{sec:distances}

\begin{table}[h]
\centering
    \resizebox{\columnwidth}{!}{
\begin{tabular}{lrrrrrrrr}
\toprule
{} &  ZH &  ES &  CS &  NWS &  EM &  RL &  TI &  FED \\
\midrule
ZH  &   \underline{.02} &                .02 &                .03 &                     .02 &              .01 &             .02 &   .05 &       .12 \\
ES  &   .03 &                 \underline{.03} &                .04 &                     .03 &              .02 &             .01 &   .06 &       .11 \\
CS  &   .02 &                .01 &                 \underline{.01} &                     .02 &              .01 &             .04 &   .06 &       .13 \\
NWS &   .05 &                .04 &                .06 &                      \underline{.04} &              .04 &             .03 &   .04 &       .09 \\
EM  &   .03 &                .03 &                .04 &                     .02 &               \underline{.03} &             .03 &   .04 &       .10 \\
RL  &   .06 &                .05 &                .07 &                     .05 &              .05 &              \underline{.05} &   .04 &       .07 \\
TI  &   .07 &                .07 &                .08 &                     .05 &              .07 &             .08 &    \underline{.02} &       .06 \\
FED &   .10 &                .10 &                .12 &                     .09 &              .10 &             .10 &   .06 &        \underline{.02} \\
\bottomrule
\end{tabular}
    }
    \caption{Wasserstein distances between the legal area distributions of the training and the test set per origin region across languages. The training sets are in the columns and the test sets in the rows.}
    \label{tab:legal_area_distribution_distances}
\end{table}

In Table \ref{tab:legal_area_distribution_distances} we show the Wasserstein distances between the legal area distributions of the training and the test sets per origin region across languages. Unfortunately, this analysis does not explain why the NWS model (zero-shot) outperforms the ZH model (in-domain) on the ZH test set, as found in Table \ref{tab:cross_domain_origin_regions}.

\section{Additional Results}
\label{sec:additional_results}

In Tables~\ref{tab:cross_domain_legal_areas_finetune}, \ref{tab:cross_domain_legal_areas_adapters},  \ref{tab:cross_domain_origin_regions_finetune} and \ref{tab:cross_domain_origin_regions_adapters} we present detailed results for all experiments. All tables include both the average score across repetitions, as reported in the original tables in the main article, but also the standard deviations across repetitions.

\begin{table*}[h]
\centering
    \resizebox{\textwidth}{!}{
    \begin{tabular}{lrcccc|c}
    \toprule
    \bf Legal Area & \bf \#D &                 \bf Public Law &                   \bf Civil Law &                    \bf Penal Law &                  \bf Social Law &                      \bf All \\
    \midrule
    Public Law & 45.6K & \underline{56.4} {\small ± 2.2} &  52.2 {\small ± 2.0} &   59.7 {\small ± 4.9} &  60.1 {\small ± 5.8} &   57.1 {\small ± 3.2} \\
    Civil Law  & 34.5K & 44.4 {\small ± 7.9} &  \underline{64.2} {\small ± 0.6} &  45.5 {\small ± 13.1} &  43.6 {\small ± 5.2} &   49.4 {\small ± 8.6} \\
    Penal Law  & 35.4K & 40.8 {\small ± 10.1} &  55.8 {\small ± 2.9} &   \textbf{\underline{84.5}} {\small ± 1.3} &  61.1 {\small ± 7.5} &  60.6 {\small ± 15.7} \\
    Social Law & 29.1K  & 52.6 {\small ± 4.2} &  56.6 {\small ± 2.0} &   69.0 {\small ± 5.5} &  \underline{70.2} {\small ± 2.0} &   62.1 {\small ± 7.6} \\
    \midrule
    \emph{All}          & 60K &   58.0 {\small ± 3.0} &  \textbf{67.2} {\small ± 1.6} &   84.4 {\small ± 0.2} &  70.2 {\small ± 1.3} &   \textbf{70.0} {\small ± 9.5} \\
    \emph{All} (w/o MT)  & 60K & 57.4 {\small ± 2.0} &  66.1 {\small ± 3.1} &   81.4 {\small ± 1.4} &  70.8 {\small ± 2.0}  &   68.9 {\small ± 8.7} \\
    \midrule
    \emph{All} (Native)  & 60K &   \textbf{58.1 {\small ± 3.0}} &  64.5 {\small ± 3.7} &   83.0 {\small ± 1.3} &  \textbf{71.1 {\small ± 4.3}} &   69.2 {\small ± 9.2} \\
    \bottomrule
    \end{tabular}
    }
    \caption{Test results for models (XLM-R with MT unless otherwise specified) \textbf{fine-tuned} per legal area (domain) or across all legal areas (domains). Best overall results are in \textbf{bold}, and in-domain are \underline{underlined}. \textbf{\emph{Cross-domain transfer is beneficial for 3 out of 4 legal areas and has the best overall results.}} The shared multilingual model trained across all languages and legal areas outperforms the baseline (monolingual BERT models). The mean and standard deviations are computed across languages per legal area and across legal areas for the right-most column. \#D is the number of training examples per legal area.}
    \label{tab:cross_domain_legal_areas_finetune}
\end{table*}

\section{Responsible NLP Research}

We include information on limitations, licensing of resources, and computing foot-print, as suggested by the newly introduced Responsible \ac{NLP}  Research checklist.

\subsection{Limitations}
\label{sec:limitations}

In this appendix, we discuss core limitations that we identify in our work and should be considered in future work.

\paragraph{Data size fluctuations}
We did not control for the sizes of the training datasets, which is why we reported them in the Tables \ref{tab:cross_domain_origin_regions}, \ref{tab:cross_domain_legal_areas} and \ref{tab:cross_jur}. This mimics a more realistic setting, where the training set size differs based on data availability. Although we discussed representational bias in RQ4, we cannot completely rule out different performance based on simply more training data.

\paragraph{Mismatch in in/out of region model performance}
As described in Section \ref{sec:cross_domain_origin_regions}, certain zero-shot evaluations outperform in-domain evaluations. Although we try to find an explanation for this in Section~\ref{sec:cross_domain_exps}, and Appendix~\ref{sec:distances}, it remains an open question since there are many confounding factors. 

\paragraph{Re-use of Indian cases}
Although we have empirical results confirming the statistically significant positive effect of training with additional translated Indian cases, we do not have a profound legal justification or even a hypothesis for this finding at the moment.

\subsection{Licensing}
\label{sec:licenses}

The \ac{SJP} dataset \cite{niklaus-etal-2021-swiss}  we mainly use in this work is available under a CC-BY-4 license. The second dataset, \ac{ILDC} \cite{malik-etal-2021-ildc}, comprising Indian cases is available upon request. The authors kindly provided their dataset.
All used software and libraries (EasyNMT, Hugging Face Transformers, deep-significance, and several other typical scientific Python libraries) are publicly available and free to use, while we always cite the original work and creators.
The artifacts (i.e., the translations and the code) we created, target academic research and are available under a CC-BY-4 license.

\subsection{Computing Infrastructure}
\label{sec:computing_infrastructure}

We used an NVIDIA GeForce RTX 3090 GPU with 24 GB memory for our experiments. 
In total, the experiments took approx. 80 GPU days, excluding the translations.
The translations took approx. 7 GPU days per language from Indian to German, French, and Italian. The translation within the Swiss corpus took approx. 4 GPU days in total.

\begin{table*}[h]
\centering
    \resizebox{\textwidth}{!}{
    \begin{tabular}{lrcccc|c}
    \toprule
    \bf Legal Area & \bf \#D &                 \bf Public Law &                   \bf Civil Law &                    \bf Penal Law &                  \bf Social Law &                      \bf All \\
    \midrule
    Public Law  & 45.6K &   \underline{57.2} {\small ± 1.8} &  53.8 {\small ± 2.1} &  58.9 {\small ± 5.2} &  61.7 {\small ± 4.1} &   57.9 {\small ± 2.9} \\
    Civil Law   & 34.5K &   41.4 {\small ± 6.6} &  \underline{57.6} {\small ± 1.1} &  42.8 {\small ± 9.1} &  43.0 {\small ± 4.1} &   46.2 {\small ± 6.6} \\
    Penal Law   & 35.4K &  37.4 {\small ± 12.8} &  56.4 {\small ± 2.0} &  \textbf{\underline{86.3}} {\small ± 0.1} &  61.6 {\small ± 6.7} &  60.4 {\small ± 17.4} \\
    Social Law  & 29.1K &   51.4 {\small ± 5.8} &  54.8 {\small ± 2.8} &  73.9 {\small ± 1.9} &  \underline{70.3} {\small ± 2.2} &   62.6 {\small ± 9.7} \\
    \midrule
    \emph{All}          &  60K &   \textbf{58.6} {\small ± 2.7} &  \textbf{66.8} {\small ± 2.8} &  83.1 {\small ± 1.3} &  \textbf{71.3} {\small ± 2.4} &   \textbf{69.9} {\small ± 8.8} \\
    \emph{All} (w/o MT) & 60K &   58.4 {\small ± 2.5} &  66.1 {\small ± 2.4} &  83.1 {\small ± 1.2} &  71.1 {\small ± 1.4} &   69.7 {\small ± 9.0} \\
    \bottomrule
    \end{tabular}
    }
    \caption{Test results for models (XLM-R with MT unless otherwise specified) \textbf{adapted} per legal area (domain) or across all legal areas (domains). Best overall results are in \textbf{bold}, and in-domain are \underline{underlined}. The mean and standard deviations are computed across languages per legal area and across legal areas for the right-most column. \#D is the number of training examples per legal area.}
    \label{tab:cross_domain_legal_areas_adapters}
\end{table*}

\clearpage

\begin{table*}[t]
    \centering
    \resizebox{\textwidth}{!}{
    \begin{tabular}{lllllllllll|l}
\toprule
\bf Region &  \bf \#D & \bf \#L &                     \bf ZH &         \bf ES &         \bf CS &    \bf NWS  &           \bf \bf EM &            \bf RL &                      \bf TI &                  \bf FED &                     \bf All \\
\midrule
ZH   & 26.4K  & de        & \underline{65.5 {\small ± 0.0}} &  65.6 {\small ± 0.0} &  63.7 {\small ± 0.0} &      68.2 {\small ± 0.0} &  62.0 {\small ± 2.9} &  57.9 {\small ± 6.7} &  63.2 {\small ± 0.0} &  54.8 {\small ± 5.1} &  62.6 {\small ± 4.1} \\
ES   & 17.1K  & de        & 62.9 {\small ± 0.0} &  \underline{66.9 {\small ± 0.0}} &  62.8 {\small ± 0.0} &      65.2 {\small ± 0.0} &  62.2 {\small ± 1.1} &  60.2 {\small ± 5.3} &  57.8 {\small ± 0.0} &  55.1 {\small ± 6.3} &  61.6 {\small ± 3.6} \\
CS   & 14.4K  & de        & 62.5 {\small ± 0.0} &  65.5 {\small ± 0.0} &  \underline{63.2 {\small ± 0.0}} &      65.1 {\small ± 0.0} &  60.7 {\small ± 1.6} &  57.8 {\small ± 3.7} &  60.5 {\small ± 0.0} &  55.9 {\small ± 0.5} &  61.4 {\small ± 3.1} \\
NWS  & 17.1K  & de        & 66.0 {\small ± 0.0} &  68.6 {\small ± 0.0} &  65.2 {\small ± 0.0} &      \underline{67.9 {\small ± 0.0}} &  61.6 {\small ± 1.7} &  57.0 {\small ± 4.9} &  57.1 {\small ± 0.0} &  55.5 {\small ± 5.7} &  62.4 {\small ± 4.9} \\
EM   & 24.9K  & de,fr     & 64.1 {\small ± 0.0} &  66.6 {\small ± 0.0} &  63.3 {\small ± 0.0} &      66.7 {\small ± 0.0} &  \underline{64.0 {\small ± 0.7}} &  66.8 {\small ± 2.9} &  63.2 {\small ± 0.0} &  58.4 {\small ± 0.3} &  64.1 {\small ± 2.6} \\
RL   & 40.2K & fr,de     & 61.0 {\small ± 0.0} &  64.7 {\small ± 0.0} &  60.2 {\small ± 0.0} &      63.7 {\small ± 0.0} &  63.4 {\small ± 3.3} &  \underline{69.8 {\small ± 2.7}} &  67.6 {\small ± 0.0} &  54.3 {\small ± 7.2} &  63.1 {\small ± 4.4} \\
TI   & 6.9K  & it        & 55.0 {\small ± 0.0} &  56.3 {\small ± 0.0} &  53.2 {\small ± 0.0} &      54.5 {\small ± 0.0} &  56.0 {\small ± 0.4} &  54.7 {\small ± 0.9} &  \underline{66.0 {\small ± 0.0}} &  53.1 {\small ± 6.4} &  56.1 {\small ± 3.9} \\
FED  & 3.9K  & de,fr,it  & 57.5 {\small ± 0.0} &  59.6 {\small ± 0.0} &  56.8 {\small ± 0.0} &      58.9 {\small ± 0.0} &  55.0 {\small ± 1.0} &  56.5 {\small ± 1.1} &  53.5 {\small ± 0.0} &  \underline{54.9 {\small ± 2.9}} &  56.6 {\small ± 1.9} \\
\midrule
\emph{All} & 60K & de,fr,it & \textbf{69.2} {\small ± 0.0} &  \textbf{72.9} {\small ± 0.0} &  68.3 {\small ± 0.0} &      \textbf{73.3} {\small ± 0.0} &  \textbf{69.9} {\small ± 1.6} &  71.7 {\small ± 2.8} &  \textbf{70.4} {\small ± 0.0} &  \textbf{65.0} {\small ± 3.9} &  \textbf{70.1} {\small ± 2.5} \\
\emph{All} (w/o MT)& 60K & de,fr,it & 68.5 {\small ± 0.0} &  71.3 {\small ± 0.0} &  67.7 {\small ± 0.0} &  71.2 {\small ± 0.0} &  69.0 {\small ± 1.5} &  71.4 {\small ± 0.3} &  67.4 {\small ± 0.0} &  64.6 {\small ± 5.2} &  68.9 {\small ± 2.2} \\
\midrule
\emph{All} (Native) & 60K & de,fr,it &  69.0 {\small ± 0.0} &  72.1 {\small ± 0.0} &  \textbf{68.6 {\small ± 0.0}} &      72.0 {\small ± 0.0} &  \textbf{69.9 {\small ± 1.6}} &  \textbf{71.9 {\small ± 0.7}} &  68.8 {\small ± 0.0} &  64.8 {\small ± 7.0} &  69.6 {\small ± 2.3} \\
\bottomrule
\end{tabular}
}
    \caption{Test results for models (XLM-R with MT unless otherwise specified) \textbf{fine-tuned} per region (domain) or across all regions (domains). Best overall results are in \textbf{bold}, and in-domain are \underline{underlined}. The mean and standard deviations are computed across languages per origin region and across origin regions for the right-most column. The regions where only one language is spoken thus show std 0. \#D is the number of training examples per origin region. \#L are the languages covered.}
    \label{tab:cross_domain_origin_regions_finetune}
    \vspace{-2mm}
\end{table*}

\begin{table*}[t]
    \centering
    \resizebox{\textwidth}{!}{
    \begin{tabular}{lllllllllll|l}
\toprule
\bf Region & \bf \#D & \bf \#L &                      \bf ZH &         \bf ES &         \bf CS &    \bf NWS  &           \bf \bf EM &            \bf RL &                      \bf TI &                  \bf FED &                     \bf All \\
\midrule
ZH   & 26.4K  & de        & 65.4 {\small ± 0.0} &  68.7 {\small ± 0.0} &  63.9 {\small ± 0.0} &      68.2 {\small ± 0.0} &  63.6 {\small ± 3.5} &  61.0 {\small ± 2.8} &  66.4 {\small ± 0.0} &  56.3 {\small ± 1.8} &  64.2 {\small ± 3.8} \\
ES   & 17.1K  & de        & 64.2 {\small ± 0.0} &  69.4 {\small ± 0.0} &  63.9 {\small ± 0.0} &      66.0 {\small ± 0.0} &  61.7 {\small ± 2.3} &  59.4 {\small ± 4.6} &  61.2 {\small ± 0.0} &  56.5 {\small ± 6.1} &  62.8 {\small ± 3.7} \\
CS   & 14.4K  & de        & 63.1 {\small ± 0.0} &  66.5 {\small ± 0.0} &  64.1 {\small ± 0.0} &      65.0 {\small ± 0.0} &  61.0 {\small ± 2.6} &  57.5 {\small ± 2.1} &  62.2 {\small ± 0.0} &  56.7 {\small ± 2.5} &  62.0 {\small ± 3.2} \\
NWS  & 17.1K  & de        & 65.8 {\small ± 0.0} &  69.0 {\small ± 0.0} &  63.8 {\small ± 0.0} &      67.4 {\small ± 0.0} &  59.9 {\small ± 3.3} &  58.6 {\small ± 1.1} &  58.9 {\small ± 0.0} &  54.2 {\small ± 2.7} &  62.2 {\small ± 4.8} \\
EM   & 24.9K  & de,fr     & 63.9 {\small ± 0.0} &  67.5 {\small ± 0.0} &  64.4 {\small ± 0.0} &      66.8 {\small ± 0.0} &  64.7 {\small ± 0.5} &  69.1 {\small ± 1.7} &  66.4 {\small ± 0.0} &  59.5 {\small ± 1.0} &  65.3 {\small ± 2.7} \\
RL   & 40.2K & fr,de     & 62.3 {\small ± 0.0} &  66.2 {\small ± 0.0} &  62.0 {\small ± 0.0} &      64.7 {\small ± 0.0} &  65.2 {\small ± 4.2} &  70.8 {\small ± 6.8} &  65.5 {\small ± 0.0} &  56.9 {\small ± 6.0} &  64.2 {\small ± 3.7} \\
TI   & 6.9K  & it        & 56.4 {\small ± 0.0} &  62.1 {\small ± 0.0} &  53.7 {\small ± 0.0} &      56.3 {\small ± 0.0} &  55.1 {\small ± 0.2} &  57.4 {\small ± 1.1} &  68.3 {\small ± 0.0} &  50.5 {\small ± 2.3} &  57.5 {\small ± 5.1} \\
FED  & 3.9K  & de,fr,it  & 52.7 {\small ± 0.0} &  52.7 {\small ± 0.0} &  51.3 {\small ± 0.0} &      53.1 {\small ± 0.0} &  52.8 {\small ± 0.7} &  52.0 {\small ± 2.3} &  52.8 {\small ± 0.0} &  50.0 {\small ± 4.0} &  52.2 {\small ± 1.0} \\
\midrule
\emph{All} & 60K & de,fr,it & \textbf{69.2} {\small ± 0.0} &  73.3 {\small ± 0.0} &  \textbf{69.9} {\small ± 0.0} &      \textbf{73.0} {\small ± 0.0} &  \textbf{70.3} {\small ± 1.9} &  \textbf{72.1} {\small ± 0.7} &  \textbf{70.9} {\small ± 0.0} &  63.8 {\small ± 6.1} &  \textbf{70.3} {\small ± 2.8} \\
\emph{All} (w/o MT)& 60K & de,fr,it & \textbf{69.2} {\small ± 0.0} &  \textbf{73.9} {\small ± 0.0} &  67.9 {\small ± 0.0} &      72.6 {\small ± 0.0} &  69.0 {\small ± 2.1} &  \textbf{72.1} {\small ± 0.3} &  70.1 {\small ± 0.0} &  \textbf{64.2} {\small ± 4.6} &  69.9 {\small ± 2.9} \\
\bottomrule
\end{tabular}
}
    \caption{Test results for models (XLM-R with MT unless otherwise specified) \textbf{adapted} per region (domain) or across all regions (domains). Best overall results are in \textbf{bold}, and in-domain are \underline{underlined}. The mean and standard deviations are computed across languages per origin region and across origin regions for the right-most column. The regions where only one language is spoken thus show std 0. \#D is the number of training examples per origin region. \#L are the languages covered.}
    \label{tab:cross_domain_origin_regions_adapters}
    \vspace{-2mm}
\end{table*}

\end{document}